\documentclass[conference]{IEEEtran}
\IEEEoverridecommandlockouts
\usepackage{cite}
\usepackage{amsmath,amssymb,amsfonts}
\usepackage{algorithmic}
\usepackage{subcaption}
\usepackage[labelformat=parens,labelsep=quad,skip=3pt]{caption}
\usepackage[export]{adjustbox}

\usepackage{graphicx}
\usepackage{textcomp}
\usepackage{xcolor}
\newcommand{\boldm}[1] {\mathversion{bold}#1\mathversion{normal}}
\usepackage[ruled,vlined,linesnumbered]{algorithm2e}
\def\BibTeX{{\rm B\kern-.05em{\sc i\kern-.025em b}\kern-.08em
		T\kern-.1667em\lower.7ex\hbox{E}\kern-.125emX}}
\begin{document}
	
	\title{Composing Distributed Data-intensive Web Services Using a Flexible Memetic Algorithm}
	
	\author{\IEEEauthorblockN{ Soheila Sadeghiram, Hui Ma, Gang Chen}
		\IEEEauthorblockA{\textit{School of Engineering and Computer Science, Victoria University of Wellington}\\
			Wellington, New Zealand  \\
			Email: Soheila.Sadeghiram $\vert$ Hui.Ma $\vert$ Aaron.Chen@ecs.vuw.ac.nz}

	}
	
	\maketitle
	\begin{abstract}
	
Web Service Composition (WSC) is a particularly promising application of Web services, where multiple individual services with specific functionalities are composed to accomplish a more complex task, which must fulfil functional requirements and optimise \textit{Quality of Service (QoS)} attributes, simultaneously. Additionally, large quantities of data, produced by technological advances, need to be exchanged between services. Data-intensive Web services, which manipulate and deal with those data, are of great interest to implement data-intensive processes, such as \textit{distributed Data-intensive Web Service Composition (DWSC)}. 
Researchers have proposed \textit{Evolutionary Computing (EC)} fully-automated WSC techniques that meet all the above factors. Some of these works employed \textit{Memetic Algorithms (MAs)} to enhance the performance of EC through increasing its exploitation ability of in searching neighbourhood area of a solution. However, those works are not efficient or effective. This paper proposes an \textit{MA}-based approach to solving the problem of distributed DWSC in an effective and efficient manner. In particular, we develop an MA that hybridises EC with a flexible local search technique incorporating \textit{distance of services}. An evaluation using benchmark datasets is carried out, comparing existing state-of-the-art methods. Results show that our proposed method has the highest quality and an acceptable execution time overall.

\end{abstract}

Web services are software modules accessible by other programs over the Web and accomplish a task \cite{channabasavaiah2003migrating}. Web services, which are provided by service providers and distributed over different locations, require some inputs and consequently generate a set of outputs after execution. Additionally, individual Web services are frequently composed together to create new Web services in order to provide some new and complex functionality. This process is called \textit{Web Service Composition (WSC)}.  Since many Web services deliver the same functionality, therefore, non-functional properties, i.e., \textit{Quality of Service (QoS)}, such as response time and cost must be considered explicitly for effective WSC. In other words, WSC is the task of selecting services from the repository and composing them together to satisfy \textit{QoS} requirements.

Moreover, data-intensive Web services deal with huge amounts of distributed data in different locations in the network and produce a huge amount of output data \cite{sadeghiram2018cluster}. These services are deal with data-intensive processes, which require performing large scale data analysis, such as indexing contents of sites or analysing large quantities of traffic logs to mine usage patterns \cite{furht2011handbook}. In \textit{Data-intensive Web Service Composition (DWSC)}, the overall quality is ensured by the careful transmission of massive quantities of data.  

 Automatically creating execution workflows while optimising the overall QoS \cite{rao2004survey}, i.e. fully-automated WSC, has attracted the attention of many researchers. Composing selected services from a large repository of Web services can be with great effort and time-consuming, as there are numerous service composition solutions to choose from within a limited time. Due to the complexity of the fully automated service composition problem, it is impossible to find optimal solutions \cite{gabrel2015web}. Significant research has been conducted on the creation of systems to compose services in an automated way, with Evolutionary Computing (EC)  \cite{aversano2006genetic,da2015graphevol,da2016memetic, da2018hybrid,da2018evolutionary, da2017fragment}, to efficiently find “good enough” composite services that meet users’ requirements reasonably well \cite{fogel2000evolutionary}. Recently,  DWSC has gained increasing interests \cite{yu2015f,yu2014hybrid,sadeghiram2018cluster}. In real-world applications, Web services are distributed over different servers. However, most of the existing approaches omit the distribution of services and data over the network, i.e. assume a centralised environment. Conversely, in a DWSC the size of data to be transferred and the location of Web services are vital in determining the overall cost and QoS of composite services.

 Many EC algorithms, such as \textit{Genetic Algorithm (GA)}\cite{holland1992genetic}, have been used for fully-automated WSC, achieving promising results\cite{da2016memetic,da2017fragment,da2018evolutionary,da2018hybrid,sadeghiram2018cluster,Sadeghiram2019Distance}. To further enhance the effectiveness of GA, researchers have developed hybridised GA enhanced with some local search techniques. This idea, which is called \textit{Memetic Algorithm (MA)} \cite{moscato1989evolution} have been successfully applied to finding high-quality solutions for WSC \cite{da2018evolutionary}. In spite of the recent success in WSC, only primitive forms of local search borrowed from algorithms designed for other relevant problems, have been applied to DWSC \cite{sadeghiram2018cluster,yu2014hybrid}. It is less effective and efficient to make local improvements without focusing on the problem itself, and without effectively utilising any information of the problem and solutions. Aside from that, defining a neighbourhood structure in performing the local search is one of the major concerns in designing memetic algorithms. Therefore, new memetic approaches must be developed to address DWSC problems addressing those issues. In this paper, we propose a new memetic algorithm for the problem of distributed DWSC in a fully-automated way, where the GA algorithm is effectively hybridised with local search techniques designed based on the location information of Web services in the distributed environment.
 
 The contributions of this paper are listed below:
 \begin{enumerate}
 
 \item To develop a novel neighbourhood-creating strategy to perform a flexible local search operator the DWSC problem considering location information of Web services. 
 \item To develop a novel memetic algorithm for the distributed DWSC. A new local search will be designed which utilises the two proposed strategy for creating the neighbourhood. 
 \item To make empirical comparisons between our proposed memetic algorithm with current state-of-the-art memetic algorithms for the DWSC and other WSC problems

\end{enumerate}



	\section{Related Works}\label{related}
Various EC techniques have been successfully applied to solving DWSC, examples are GA for WSC \cite{da2018evolutionary,da2016memetic,canfora2005approach}, Genetic Programming (GP)\cite{koza1992genetic} for WSC \cite{da2016memetic,da2018evolutionary} and Particle Swarm Optimisation (PSO) \cite{da2016particle}. The main difference between GA and GP is that GA uses sequences to represent chromosomes (solutions), while in GP, solutions are represented as trees. The application of genetic operators on sequences in GA is more straightforward than trees in GP, i.e., operators can be applied without restrictions since the functional correctness of the composition will be ensured during decoding. Additionally, the superiority of the GA-based method to Particle Swarm Optimisation \cite{kennedy2011particle} and integer linear programming methods are demonstrated in \cite{da2018evolutionary} and \cite{canfora2005approach}, respectively.
A group of approaches have investigated fully-automated DWSC \cite{yu2015f,yu2014hybrid,sadeghiram2018cluster}.
A hybrid approach combining GP and Tabu Search \cite{glover1989tabu} is proposed to solve the
DWSC problem \cite{yu2014hybrid} with centralised services. However, this method does not clarify how it checks the
validity of a solution. Another approach to DWSC is proposed in \cite{yu2015f}, in which tree-based representation and search space reduction techniques are adopted before the optimisation starts using fully and partially dependent Web services.

 In particular, when global optima are located in immediate vicinities of some existing solutions, it might take a significantly longer time to identify optima. On the other hand, emphasising the synergy between exploration and exploitation, MA can exploit the neighbourhood area of an individual well and obtain sufficient precision with the help of local search at the expense of execution time. In the literature, MAs have been shown to be extremely effective for WSC \cite{da2016memetic,da2018evolutionary,sadeghiram2018cluster, yan2016evolutionary,Sadeghiram2019Distance}.
An MA based on GraphEvol \cite{da2015graphevol} method has been suggested for WSC problem \cite{yan2016evolutionary}, in which, two move operators are designed for locally modifying an individual on the graph while keeping its feasibility. One operator focuses on the service selection, and the other considers modifying the graph structure. 
MA with indirect representation for WSC has been first introduced in \cite{da2016memetic}. In that paper, a problem specific crossover has been designed and a swap local search has been applied to the chromosomes of GA. The indirect representation reduced the overall execution time while maintaining the original solution quality. Additionally, the use of the memetic local search improved the overall quality of solutions.

However, most of existing WSC approaches assume a centralised service repository for the composition, which is unrealistic for WSC and even more unsuitable for DWSC where the location of data and Web services is a fundamental element. 
Only a small number approaches have addressed the composition problem considering the distributed nature of Web services \cite{sadeghiram2018cluster,Sadeghiram2019Distance}. Indirect representation was successfully applied to distributed DWSC \cite{sadeghiram2018cluster}, and the initial population of GA has been created with the help of clustering, where Web services have been grouped into clusters based on their relative geographical distances. Clusters were then combined together using a crossover operator. Additionally, the effectiveness of considering distance in distributed DWSC has been verified in \cite{sadeghiram2018cluster}, where the distance between services has been applied as a heuristic in creating the initial population for GA with fixed-length chromosomes. In another research, the distance between services has been applied to design crossover and local search operators for MA\cite{Sadeghiram2019Distance}.

\section{Problem Definition and Objective Function}\label{defination}

In this section, first the definition of the distributed DWSC based on the definition in our previous paper \cite{sadeghiram2018cluster} will be presented, which includes some basic terms for the distributed DWSC problem, and afterwards, the objective function of the problem will be presented.

\subsection{Basic Concepts and Terminology}\label{problemDes}
First, we define basic concepts that should be used later in the objective function. A \textit{data-intensive Web service} is a tuple \textit{$S_{i}$= ( $I_{i}$, $O_{i}$, $QoS_{i}$, $D_{i}$, $l_{i}$)}, where {\textit{$I_{i}$}} is a set of inputs. $S_{i}$ is the $i_{th}$ service in a repository $\mathcal{R}$ and $O_{i}$ is a set of outputs of the service $S_{i}$. $QoS_{i}$ is the set of quality attributes of the service which describes non-functional properties. In this paper, for each Web service, we consider $T_{i}$ and $C_{i}$, which refer to the total time and cost required for executing service $S_{i}$. $D_{i}$ is the set of $m$ data items $d_{j}$, $ j \in \{1,...,m\}$ required by service $S_{i}$ and $l_{i}$ is the location of $S_{i}$.

A \textit{service repository} $\mathcal{R}$ consists of a finite collection of Web services $S_{i}$, $ i \in \{1,...,n\}$. A \textit{service request} (also called a \textit{composition task}) is a tuple $\mathcal{T}=(I_\mathcal{T}, O_\mathcal{T})$ where $I_\mathcal{T}$ is a set of task inputs a user can provide for the composition, and $O_\mathcal{T}$ is a set of task outputs expected by the user to be produced by the composition.

\textit{Data} is a tuple \textit{ d= ($cost_{d}$, $size_{d}$, $l_{d}$ )}
where \textit{$cost_{d}$} is the cost applied by the data provider, i.e., the cost to provide data, \textit{$size_{d}$} is the data size, and \textit{$l_{d}$} is the location of the data centre that hosts the data.

For a given \textit{Task}, we need to find a \textit{composition} that fulfils the request. A service composition is often represented as a Directed Acyclic Graph (DAG) which includes a set of services that could jointly accomplish the required task, where two special services can be used to represent the overall composition's inputs and outputs: a start service $S_0$ with $I(S_0)$ = $\varnothing$ and $O(S_0)$ = $I_\mathcal{T}$, and an end service $S_{n+1}$ with $I(S_{n+1})=O_\mathcal{T}$ and $O(S_{n+1})$= $\varnothing$. In a composite Web service, there is a communication link between $S$ and $S'$ if some outputs of service $S$ are used as inputs for $S'$. In this paper, composite services can support both parallel and sequence constructs. The parallel construct allows services to be executed in parallel, meaning that their inputs are fulfilled independently and consequently their outputs are produced independently from each other. We need to ensure that composite services are $feasible$. In particular, we need to ensure that a Web service can directly connect to another only when its outputs can satisfy at least one of the other services input(s).

\subsection{Quality of Composite Services and the Objective Function}
 A significant amount of time is spent on transferring and accessing data during the execution of a composite data-intensive Web service in a distributed environment. An example of a composite service for DWSC is illustrated in Fig. 1. For simplicity, all associated time and cost values are shown only for one Web service, one connection link and one data. The following definitions have been used to accurately capture the various time and cost components involved \cite{sadeghiram2018cluster}. 
 \begin{figure}
\includegraphics[ width=0.45\textwidth]{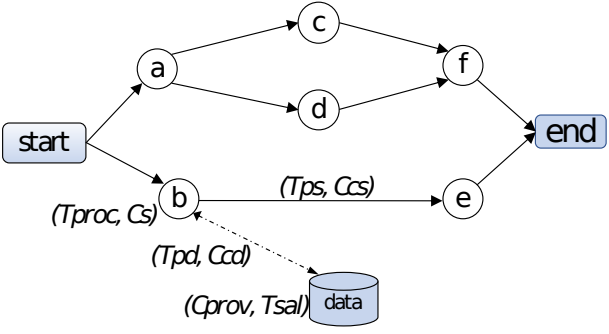}
\label{exampleofdefinitions}
\caption{A composite service and its components }
 \end{figure}

	\textit{Server access latency \begin{math}({Tsal})\end{math}}: the time to retrieve data from a data centre.
	
	\textit{Data execution time \begin{math} ({Tproc})\end{math}}: time for processing a data and executing the service.
	
	\textit{Service cost \begin{math} ({Cs})\end{math}}: the cost to use the service.
	
	\textit{Data provision cost \begin{math} ({Cprov})\end{math}}: the cost applied by a data provider, i.e., the cost to use the data.
	
	\textit{Data transfer time \begin{math} ({Tt})\end{math}}: the amount of time that it takes for a server to put all bits of the given data over the communication link and is defined as the data size divided by the network bandwidth.

		\textit{Propagation delay \begin{math} ({Tp})\end{math}}: the time it takes for the data to travel from the sender to the receiver. The sender is either a server hosting the data or a server hosting a preceding service in the composition which produces required data for the receiver. In the first case, we denote the propagation time as \begin{math} {Tpd}\end{math} while in the second case we denote it as \begin{math} {Tps}\end{math}. In both cases, the receiver is a server hosting a Web service which takes that data as its input.

	\textit{Communication cost \begin{math} ({Cc})\end{math}}: the cost to transfer data from one Web service to another. We denote it as \begin{math} {Ccs}\end{math} if the data is transferred between two Web services, while we use \begin{math} {Ccd}\end{math}  to represent the cost to move data from the hosting server to a Web service.

Correspondingly, the total execution time and cost of a Web service $S_{i}$, i.e., $T_{i}$ and $C_{i}$, (including data-related time and cost) are calculated in Equations \eqref{tatomic} and \eqref{catomic}, respectively.

\begin{equation}
T_{i}={\sum_{j=1}^{m}(Tpd_{d_{j}}+Tsal_{d_{j}}+Tproc_{d_{j}}+Tt)}
\label{tatomic}
\end{equation}

\begin{equation}
C_{i}={\sum_{j=1}^{m}(Ccd_{d_{j}}+Cd_{d_{j}}+C_s})
\label{catomic}
\end{equation}

In the above functions, $m$ is the total number of data items in $D_{i}$.

The overall cost is obtained by summing up the costs of all services in the composition, i.e., nodes (services) and associated costs for edges (communication links) in the graph, as shown in Equation \eqref{ci}:

\begin{equation}
C_{total}=\sum_{i=1}^{NODE} C_{i}+\sum_{i=1}^{EDGE}{Ccs_{i}}
\label{ci}
\end{equation}
\noindent \begin{math} {Ccs} 
\end{math} is the communication cost. \begin{math} NODE \end{math} and \begin{math} EDGE \end{math} are the total numbers of nodes (Web services) and edges (links between services) included in that composition, respectively. 

Response time $T_{total}$ is the time of the most time-consuming path in the composition. Assuming $h$ is the number of paths in a composite service,
\begin{math} N{p} \end{math} and \begin{math} E{p} \end{math} are the number of nodes and edges in a path $p$, respectively. The overall time is defined as in Equation \eqref{ti}:  
\begin{equation}
T_{total}=\max_{p=1}^h (\sum_{i=1}^{N{p}}{T_{i}}+\sum_{i=1}^{E{p}}{Tps_{i}})
\label{ti}
\end{equation}

Finally, the goal is to minimise the function in Equation \eqref{fitness} subject to the compositions in the set of all composite services over a given repository of individual services. Accordingly, the best solution will be a composition with the minimum value of $F$ which function will be used as the fitness measure in our MA algorithm. 
\begin{equation}
F=w_{t}\hat{T}_{total}+w_{c}\hat{C}_{total}
\label {fitness}
\end{equation}
where \begin{math}
\hat{T}_{total}\end{math} and \begin{math}\hat{C}_{total} \end{math} are normalised values of \begin{math}{T}_{total}\end{math} and \begin{math}{C}_{total} \end{math}, respectively. \begin{math}w_{t} \end{math} and \begin{math} w_{c} \end{math} are positive real weights, defined in the task and $w_{t}+w_{c}=1$. The upper bound for normalisation is calculated as in \cite{da2018evolutionary}.

In this section, we introduced the distributed DWSC problem. We addressed the problem of current research, i.e., we considered the distribution of data and services over different locations, therefore, communication time and cost are vital to the overall quality of any composite services and must be addressed explicitly. 
\section{Memetic Algorithm for DWSC}\label{design}
 The dual concept of exploitation and exploration consists of two fundamental aspects complementing any effective search method. Memetic Algorithms (MAs) offer mechanisms to achieve this general objective. Literally, MAs are hybrid search methods based on a population-oriented search strategy, e.g., GA, and a local search technique where their success is often attributed to the algorithms' ability to properly combine explorative and explorative searches. For example, GA locates the regions where the global optimum exists and the local search helps to converge quickly to the optimum.
 
This paper focuses on the neighbourhood structure, where a solution's neighbourhood will be defined by adding some related extra services to the original solutions chromosome. Relative distances of Web services, which is a key criterion for communication cost and time, will be utilised to identify where to insert these extra services in the solution's chromosome. That is to say, we will use domain knowledge of DWSC to produce more flexible local search operator and propose an effective and efficient MA algorithm, compared to the state-of-the-art local search technique \cite{da2018evolutionary,Sadeghiram2019Distance}, for the distributed DWSC problem. 

Additionally, the MA relies on a crossover operator to combine parts of different individuals, to derive new solutions. New individuals may be modified by other operators, such as mutation, before being added to the population. In this paper, we adopt the same mutation operator \cite{da2018evolutionary}, due to its proven effectiveness. The pseudocode of the MA for DWSC is shown in Algorithm \ref{algorithm1}. In particular, the relative distance of services plays an important role in the local search and crossover operators. Crossover operator and local search will be discussed in \ref{operators} and \ref{loc}, respectively.

We use an indirect representation, i.e. sequences,  with variable-length chromosomes, to allow flexibility in designing EC operators \cite{Sadeghiram2019Distance}. The initial population is created by randomly ordering all the services in the repository in sequences. All EC operators are applied to sequences. Especifically, the algorithm and operators must be able to maintain the feasibility of solutions in terms of functionality constraints while improving time and cost properties. This representation was firstly proposed for WSC in \cite{da2016particle} and then successfully utilised by other relevant works \cite{sadeghiram2018cluster,da2016memetic,da2018evolutionary}.

\begin{algorithm}[!htb]
	\setlength\hsize{0.9\linewidth}
	\SetKwInOut{Input}{Input}\SetKwInOut{Output}{Output}
	\SetKwFunction{filterByLayer}{filterByLayer}\SetKwFunction{mergeLayers}{mergeLayers}\SetKwFunction{findHighestTime}{findHighestTime}
	\SetKwFunction{eNull}{null}\SetKwFunction{getInputsSatisfied}{getInputsSatisfied}\SetKwFunction{calculateFitness}{calculateFitness}
	\LinesNumbered
	\SetNlSty{}{}{:}
	\Input{$Service~Repository$ ($\mathcal{R}$ )}
	\Output{$A~Service~Composition~Solution$ }
	
	Generate sequences with randomly ordered services in $\mathcal{R}$;
	
	Decode sequences and calculate the fitness of each sequence;
	
	Update sequences by removing redundant services not used during the decoding;

	\While{the number of iterations not reached}{
		Select sequences based on their fitness value using a tournament selection;
		
		Apply crossover operator;
		
		Apply mutation operator;	
		
		Apply local search (both Type-I and Type-II ) to the $current~sequence$ to produce neighbour solutions;
		
		}
	
	\Return $SequenceWithBestFitness$\;
	\caption{Memetic algorithm for DWSC}
	\label{algorithm1}
\end{algorithm}

\subsection{Representation of Chromosomes}

 In this paper, we use indirect representation, in the form of sequences of services, which allows the optimisation to be carried out without any restrictions since functional constraints are subsequently enforced during the decoding step. A decoding algorithm transforms sequences into a corresponding executable service composition, i.e., a feasible workflow \cite{da2018evolutionary,sadeghiram2018cluster}. In fact, the decoding process generates workflows automatically which is the requirement of a fully-automated DWSC approach. An example of backward decoding of a sequence (the solution is made from the end service, $S_{n+1}$ to the start service $S_0$ ), and the sequence after being decoded are illustrated in Fig. 2. Redundant services, which have not been used in the solution, are removed from the sequence in a variable-length GA. That is to say, during the decoding each candidate is reduced to contain only those services used in the composition solution. In addition, our EC operators will produce chromosomes with duplicated services that might affect the efficiency of the algorithm. Therefore, duplicated services are removed from the sequence before the decoding starts. 

\begin{figure}

	\label{fig:backdecode}
	
		\centering
	\begin{tabular}{@{}c@{}}
		

			\includegraphics[width=8.5cm, height=5cm]{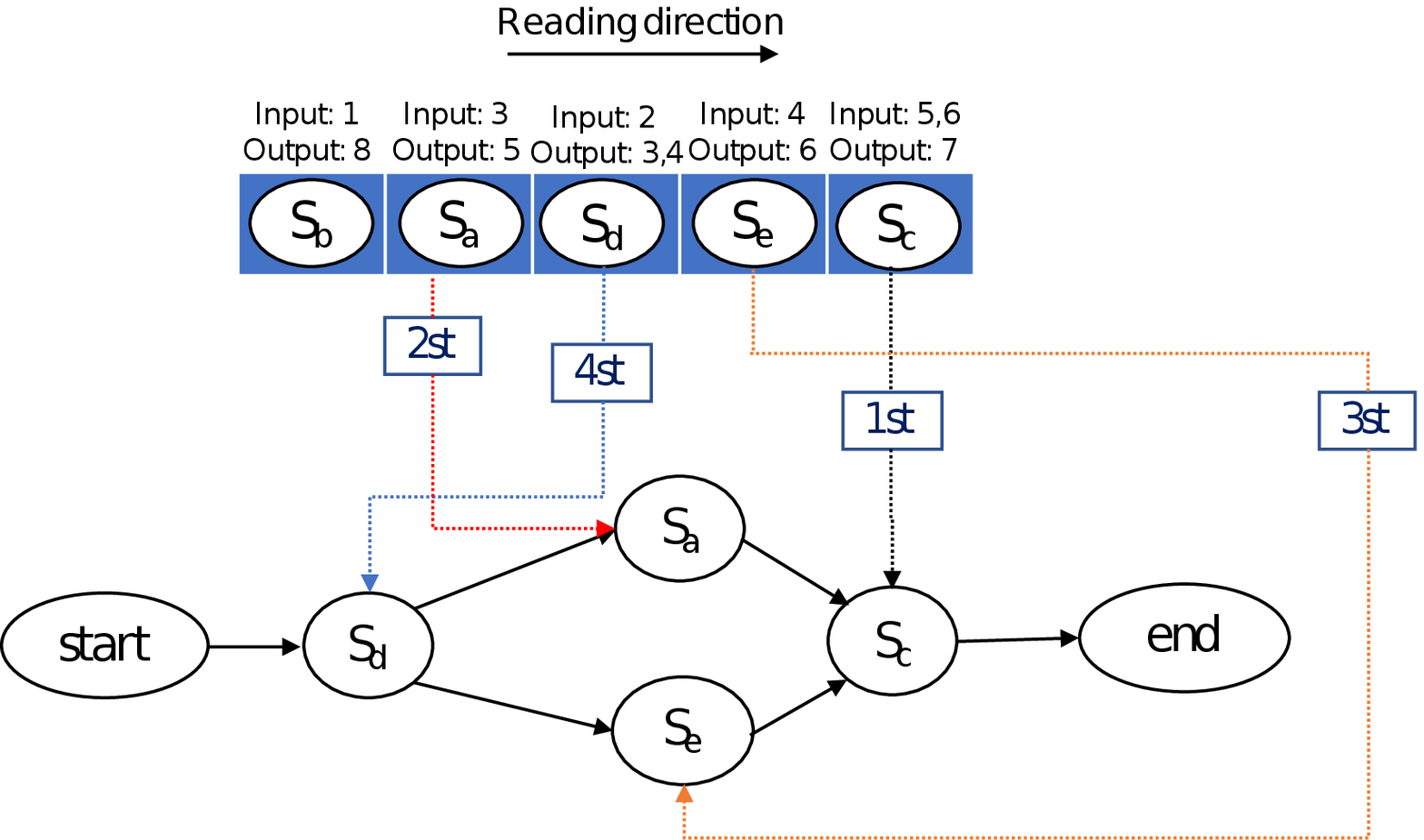}\\[\abovecaptionskip]
		\label{fig:backward}
				\small  (a) 
	\end{tabular}
	
	
	\begin{tabular}{@{}c@{}}
		\captionsetup{justification=centering}


	\includegraphics[width=4cm]{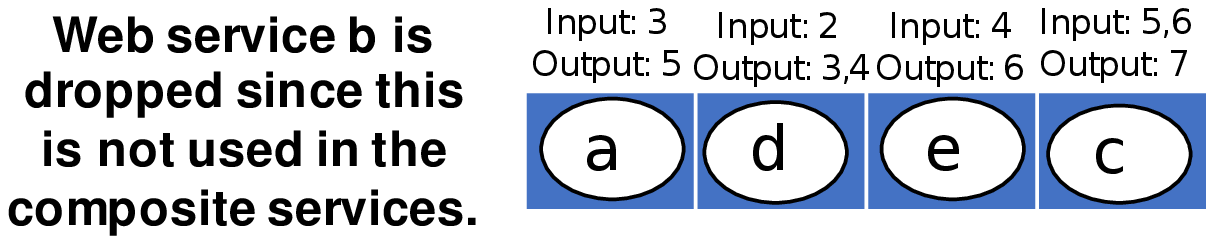}\\[\abovecaptionskip]
			\small (b) 
		\label{fig:afterdecod}
	\end{tabular}

			\caption{(a) Backward decoding example (note that the sequence is traversed as many times as possible), (b) the sequence after being decoded in a variable-length GA.}
\end{figure}

\subsection{Crossover} \label{operators}
\begin{figure}
	\centering
	\includegraphics[width=9cm]{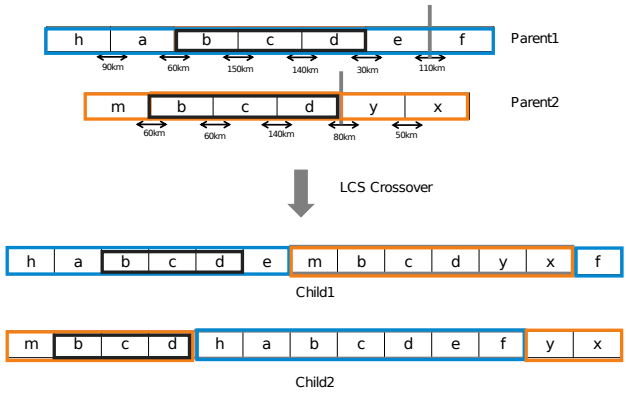}
	\caption{Example of distance-guided LCS crossover for the proposed method. The longest common subsequence is identified by an orange rectangle.}
	\label{fig:LCS crossover}
\end{figure}

In this paper, we apply distance-guided LCS crossover. This crossover which was designed to incorporate domain knowledge (e.g. location of services) into GA, was introduced in \cite{Sadeghiram2019Distance}. This crossover outperformed three other crossovers which were developed for WSC. Distance-guided crossover is illustrated in Fig. \ref{fig:LCS crossover}, where a heuristic, called the longest common subsequence (LCS), is incorporated to preserve the promising part of each parent and transfer it to children. This heuristic finds the longest common subsequence between two parents, i.e., the longest sequence of services which appears in both parents. The crossover point then is set immediately after that Web service  with longest communication link, but never appears directly within the LCS. For more information about distance-guided LCS crossover, refer to \cite{Sadeghiram2019Distance}.

\subsection{Local Search}\label{loc}
One major difficulty in designing local search for MAs lays in the definition of neighbourhood structure \cite{yan2015adaptive}. In this paper, the aim of local search is to avoid long communication links, or bottlenecks, which lead to an increase in the overall communication time and cost. We will find the bottleneck after a solution is generated, and then, changes will be applied to the place of this link in the corresponding sequence to produce neighbour solutions. This paper introduces two local search strategies to build an effective MA for DWSC by incorporating communication information. 

\begin{figure}
	\begin{subfigure}{6cm}
\includegraphics[width=8cm]{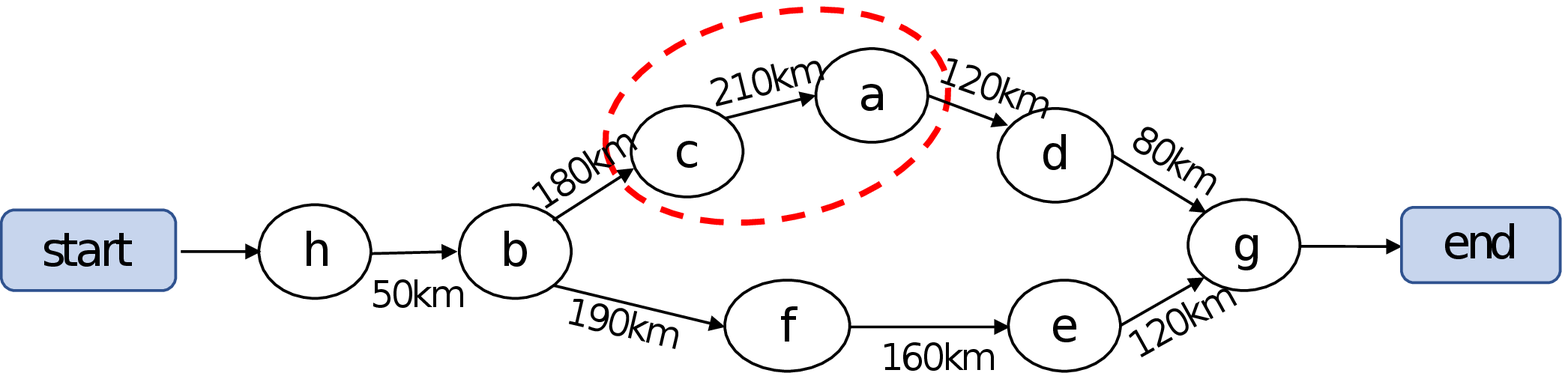}
\caption{}
\label{fig:comp}
\captionsetup{justification=centering}
	\end{subfigure}
\begin{subfigure}{6cm}
\includegraphics[width=9cm]{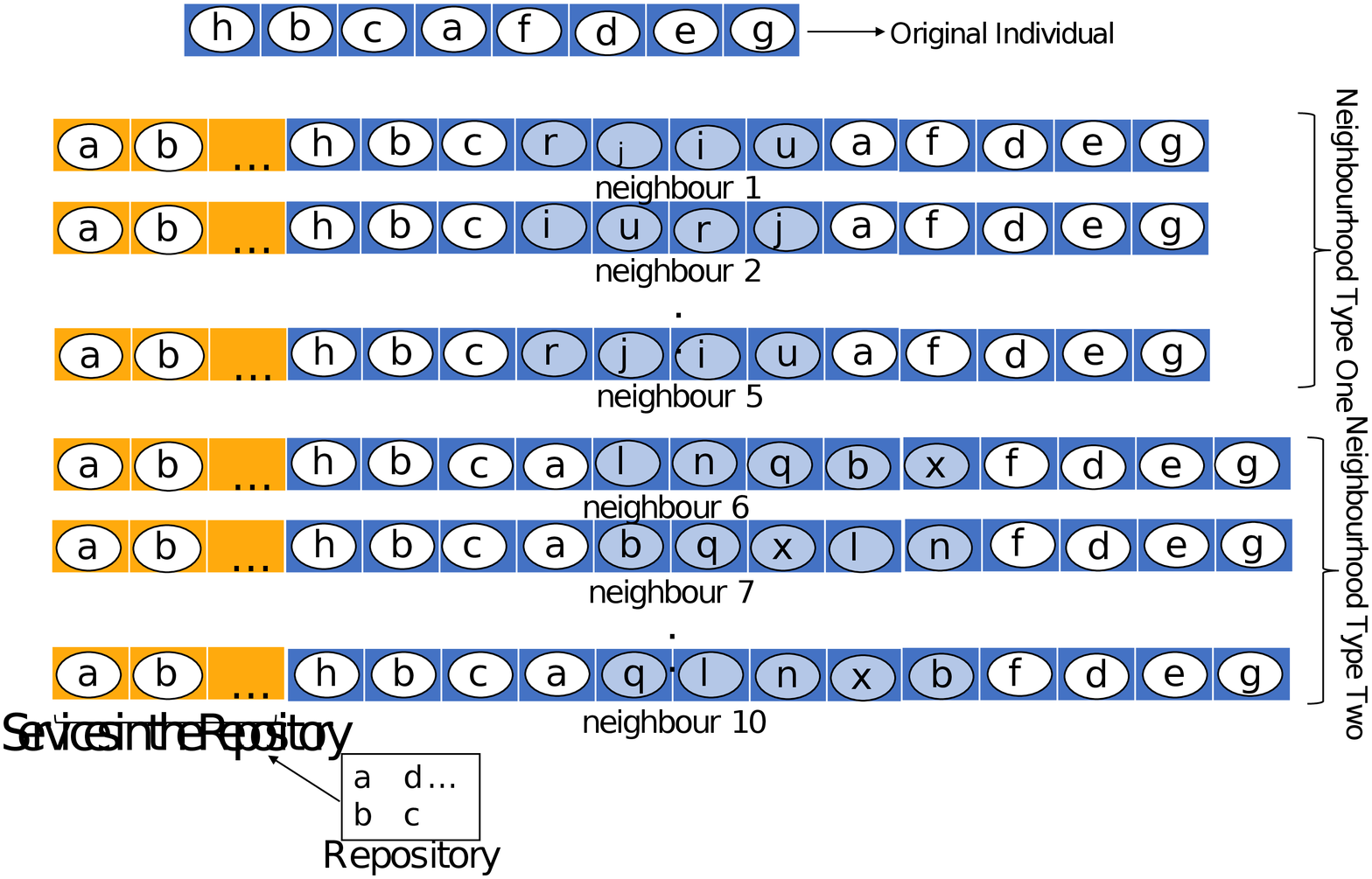}
\caption{}
\captionsetup{justification=centering}

\label{fig:LS}
\end{subfigure}
\label{fig:Localsearch}
\caption{(a) A solution example and a bottleneck link, (b) original sequence and two neighbourhoods for local search. }
\end{figure}

 To demonstrate the proposed local search strategies an example of a solution is illustrated in Fig. \ref{fig:comp}, where, the longest communication link of the solution and its two corresponding services, i.e., Web services $c$ and $a$, are shown inside the dashed area. This link is considered as a bottleneck because the communication time and cost are defined based on the distance between services. Therefore, if we can escape this link or replace it and the two services by another alternative path that performing the same task but with a shorter communication link, we will possibly enhance the total performance of the solution. 

The first strategy in designing local search for the MA is to create neighbours through replacing the Web service $a$ by another alternative service that its input(s) can connect to the output(s) of $c$, assuming that at least some of those services are probably located in a shorter distance to $a$. After finding all of those alternative services from the repository, we insert them after the Web service $c$ in the corresponding sequence. We call this local search Type-I, and it is indicated in Fig. 4(b). Random orders can be followed while inserting these services in order to produce different new neighbours.  Moreover, a prefix including random services from the repository (except those in the sequence) will be attached to the beginning of neighbour sequences. This is important for the purpose of diversifying the composite services in the neighbourhood. Redundant services of the sequence which would not be part of the composition graph will be eliminated during the decoding stage.

Referring to Fig. 4(a) as an example, the second strategy is to consider alternatives for the whole part identified within the red dashed area, which includes $a$, $c$ and their communication link. In other words, we do not have to maintain the Web service $c$ which forces us to consider only feasible services that can follow immediately after service $c$. The justification of this local search is that if $c$ itself is located on a very distant server, (which is most likely), solution quality can be further improved by avoiding using it. Therefore, in this local search, which we will call it Type-II, we place those services in the sequence just after $f$. The idea behind this local search is to replace a relatively larger part of a solution rather than a single Web service. In this way, we introduce more search capability to the local search operator in finding good solutions in the neighbourhood. In comparison, local search Type-I, which only considers one Web service replacement, is more restrictive than Type-II. 

Same as the first strategy, the order of inserted services will be shuffled for creating a new neighbour, and a subsequence of remaining services in the repository will be attached to the beginning of the solution. 

Both strategies will be used in the local search utilised in the memetic algorithm for DWSC. The pseudocode of this local search procedure is shown in Algorithm \ref{algorithmLS}. A solution's neighbours are created by local search Type-I and Type-II. Afterwards, they are evaluated and the original solution is replaced by its best neighbour. If there are no improvements in neighbours, no replacement will be performed.

\begin{algorithm}[!htb]
	\setlength\hsize{0.9\linewidth}
	\SetKwInOut{Input}{Input}\SetKwInOut{Output}{Output}
	\SetKwFunction{filterByLayer}{filterByLayer}\SetKwFunction{mergeLayers}{mergeLayers}\SetKwFunction{findHighestTime}{findHighestTime}
	\SetKwFunction{eNull}{null}\SetKwFunction{getInputsSatisfied}{getInputsSatisfied}\SetKwFunction{calculateFitness}{calculateFitness}
	\LinesNumbered
	\SetNlSty{}{}{:}
	\Input{$Current~Solution$, $ N_l(Neighbourhood Size)$, $Service~Repository$ }
	\Output{$Neighbouring~Solution$ }
	
	Find two consecutive services with the longest distance in the composition solution;

	Generate $N_l$ neighbours for $Current~Solution$, where $N_l$/2 is obtained using the local search Type-I, and $N_l$/2 is obtained using the local search Type-II;
	
	Evaluate all neighbours; 

	\If{$Best~Neighbour$'s fitness is better than $Current~Solution$'s fitness}
	 {$Current~Solution$=$Best~Neighbour$;}

	\Return $Current~Solution$\;
	\caption{ Local search for DWSC}
	\label{algorithmLS}
\end{algorithm}

\section{Experiment Design}\label{ex}

To examine the performance of the proposed MA we conduct experiments using WSC-2008 \cite{bansal2008wsc} and WSC-2009 \cite{kona2009wsc} benchmark datasets. WSC-2008 contains eight service repositories of varying sizes, while WSC-2009 includes five repositories with a greater variety of sizes. A taxonomy of concepts is provided for determining the matching relationship between any pair of services, and a number of service composition tasks are also given \cite{bansal2008wsc,kona2009wsc}. These datasets were chosen because they are the largest and most frequently used benchmarks; however, they do not contain the location information of servers which hosts Web services and data. For our purpose, the distance between two Web services is estimated using the same method as proposed in \cite{sadeghiram2018cluster} based on the coordinates information in WS-dream open dataset \cite{zheng2014investigating}.

Random values that are drawn from a normal distribution are assigned to the netwrok bandwidth, which is utilised in calculating \begin{math} {Tt}\end{math}, for each connection link in the interval (0,1]. 
Additionally, each data has has its own \begin{math} {Tsal}\end{math} which relates to the server and a and \begin{math} {Cprov}\end{math}, which are both generated randomly in the interval (0,1].
Data size has been considered same, i.e. equal to 3, for all data. Therefore, values of \begin{math} {Cc}\end{math}, including \begin{math} {Ccs}\end{math} and \begin{math} {Ccd} \end{math}, and \begin{math} {Tp} \end{math} only depend on the distance between services, and are drawend from the dataset WS-dream.

Values of \begin{math} {Tproc}\end{math} and \begin{math} {Cs}\end{math} are  obtained from datasets WSC-2008 and WCC-2009, and then normalised to fit between 0 and 1. 

Values of weight parameters are \begin{math} w_{t}=w_{c}=0.5 \end{math}, which means that the time and cost have identical contributions to the fitness. Finally, the final fitness value will fit between 0 and 1 since \begin{math} w_{t}+w_{c}=1 \end{math}. 

Our MA for DWSC has been compared with two other techniques, where both of them are adapted to use the same crossover and mutation as proposed MA. The first approach to be compared with is drawn from MA for WSC \cite{da2018evolutionary} but it is adapted for DWSC and applies LCS local search (MA-WSC). The second approach is MA for DWSC proposed in \cite{Sadeghiram2019Distance}, i.e., MA-DWSC(I). Each algorithm has been run for 30 independent times on desktop computers with 8 GB RAM and an Intel Core i7-4790 processor (3.6GHz). 
Along with crossover and mutation operators, all methods have the same parameter set in order to have a fair comparison. The numer of generations is 100. Local search, mutation and crossover probabilities are 0.05, 0.05 and 0.95, respectively. These parameters' values were chosen based on popular settings discussed in the literature \cite{koza1992genetic,shi2001particle}.
Tournament selection of size two is adopted as the strategy for choosing which candidates to update in these approaches. It was also utilised to select individuals to undergo EC operators, i.e. mutation, local search and crossover.
 Neighbourhood size for local search is 20 for our method, which we call it MA-DWSC(II) and MA-DWSC(I), however, MA-WSC does not have a fixed neighbourhood size because it depends on the length of the sequences, i.e., the number of services in the sequence. Elitism size is two for all methods. 

\subsection{Results}

Table \ref{tab:results} shows the mean solution fitness and standard deviation for the 30 independent runs of each approach. MA-DWSC(I) only uses local search Type-I \cite{Sadeghiram2019Distance}, however, in MA-DWSC(II) we produced neighbours through both local search strategies, i.e., Type-I and Type-II. MA-WSC uses the local search technique developed in \cite{da2018evolutionary}. To verify whether fitness values for each approach were significantly different, analysis of variance statistical test (ANOVA) at 0.05 significance level has been conducted. As illustrated in Table \ref{tab:results}, for task 08-1 the fitness achieved by the MA-DWSC(II) was significantly better than the two other techniques. On the other hand, for task 08-3 there was no significant difference among all algorithms 
Moreover, results show that the quality of solutions which are produced using MA-DWSC(II) are generally better than the results of the two other methods. The reason can be explained in the local search operator (since all methods are using same mutation and LCS crossover). MA-DWSC(I) and MA-DWSC(II) use problem-specific local search operators. However, local search in MA-DWSC(I) is more restrictive because it only targets one Web service. On the other hand, MA-DWSC(II) has achieved good results due to its use of flexible operators, which generates neighbours by replacing a path.

\begin{table}
	\centering
	\caption{Mean fitness values and standard deviations per 30 runs. The significantly better values are shown in bold for each task. (Note: the lower the fitness the better)}
	\label{tab:results}
	\begin{center}

		\begin{tabular}{|l|l|l|l|}

			\hline
			\multicolumn{1}{|p{1.4cm}|}{\centering Task} & 
			\multicolumn{1}{|p{1.3cm}|}{\centering MA-WSC} &  \multicolumn{1}{|p{2.2cm}|}{\centering MA-DWSC(I)\cite{Sadeghiram2019Distance}} & \multicolumn{1}{|p{1.8cm}|}{\centering MA-DWSC(II)}     \\

			\hline

			WSC08-1  &{ $0.41\pm0.12$}  & { $0.42\pm0.02$} & { \boldm$0.4\pm0.04$} \\
			
			WSC08-2  & { $0.42\pm0.02$} & $ 0.46\pm0.04$ & {\boldm $0.41\pm0.02$} \\
			
			WSC08-3  & { $0.44\pm0.02$}&  { $0.47\pm0.02$}& { $0.44\pm0.03$}  \\
			
			WSC08-4  & { $0.4\pm0.01$}& { $ 0.41\pm0.01$}& { \boldm $0.39\pm0.02$}  \\
			
			WSC08-5   & { $0.46\pm0.16$} &  { $ 0.48\pm0.04$}& { \boldm $0.45\pm0.02$}
			\\
			
			WSC08-6  & { $0.46\pm0.22$}& { $0.51\pm0.15$}& { $0.47\pm0.02$} \\
			
			WSC08-7  & { $0.53\pm0.02$}& { $0.56\pm0.02$}& { \boldm $0.52\pm0.02$}     \\
			
			WSC08-8 & {\boldm $0.45\pm0.05$} &  {\begin{math} 0.48\pm0.08\end{math}} & { $0.46\pm0.09$} \\
			
			WSC09-1 & { $0.53\pm0.02$} & {{ $ 0.56\pm0.05$}} & { \boldm$0.51\pm0.02$}\\
			
			WSC09-2  & { $0.47\pm0.02$} & \begin{math} 0.5\pm0.01\end{math} & { \boldm$0.45\pm0.22$} \\
			
			WSC09-3 & { $0.499\pm0.06$}& { $ 0.523\pm0.001$}& {\boldm $0.48\pm0.1$}  \\
			
			WSC09-4  & { \boldm$0.47\pm0.09$} & {  $0.51\pm0.21$} & { \boldm $0.48\pm0.02$} \\
			WSC09-5  & {\boldm $0.42\pm0.03$}& {$0.48\pm0.06$}   & { $0.46\pm0.06$}  \\
			\hline
			
		\end{tabular}%
	\end{center}
	
\end{table}

Similarly, mean execution time(s) for each approach and ANOVA statistical test results are indicated in Table \ref{tab:time}.

According to Table \ref{tab:results} on the tasks 08-8, 09-4 and 09-5 MA-WSC produced satisfactory results. This is probably due to the large neighbourhood. Neighbourhood size in this method is not fixed but depends on the number of Web services in the sequence (unlike the two other methods). For example, for task 09-5 the average neighbourhood size for 30 runs is 90. On the other hand, a large neighbourhood size can cause an increase in the execution time as it is clearly shown in Table \ref{tab:time}. 

\begin{table}
	\centering
	\caption{Mean execution time for each approach per 30 runs. The significantly lowest times are shown in bold for each task.}
	\label{tab:time}
	\begin{center}

		\begin{tabular}{|l|l|l|l|}

			\hline
			\multicolumn{1}{|p{1.4cm}|}{\centering Task} & 
			\multicolumn{1}{|p{1.3cm}|}{\centering MA-WSC} &  \multicolumn{1}{|p{2.2cm}|}{\centering MA-DWSC(I)\cite{Sadeghiram2019Distance}} & \multicolumn{1}{|p{1.8cm}|}{\centering MA-DWSC(II)}     \\

			\hline

			WSC08-1  &{ $0.5\pm0.06$}  & {$0.45\pm0.12$} & { $0.43\pm0.03$} \\
			
			WSC08-2  & { $0.57\pm0.13$} & {$ 0.42\pm0.14$} & {\boldm $0.4\pm0.02$} \\
			
			WSC08-3  & { $1.32\pm0.07$}&  { $1.14\pm0.42$}& { \boldm$1.28\pm0.02$}  \\
			
			WSC08-4  & {\boldm $0.7\pm0.02$}& { $ 0.71\pm0.01$}& { $0.74\pm0.02$}  \\
			
			WSC08-5   & { $0.98\pm0.11$} &  { $ 1.11\pm0.04$}& { $1.02\pm0.36$}
			\\
			
			WSC08-6  & { $5.52\pm0.49$}&   { $4.91\pm0.15$}& { \boldm $4.45\pm0.17$} \\
			
			WSC08-7  & { $2.64\pm0.23$}& {\boldm $2.16\pm0.02$}& { $2.42\pm0.19$}     \\
			
			WSC08-8 & { $5.68\pm0.48$} &  { $5.32\pm0.31$} & { \boldm $5.18\pm0.12$} \\
			
			WSC09-1 & { $0.46\pm0.01$} &  {{ $ 0.45\pm0.02$}} & { \boldm $0.44\pm0.01$}\\
			
			WSC09-2  & {$4.18\pm0.32$} & {\boldm$ 4.02\pm0.01$} & { \boldm $4.02\pm0.02$} \\
			
			WSC09-3 & { $4.32\pm0.2$}& { $ 4.32\pm0.001$}& { $4.09\pm0.02$}  \\
			
			WSC09-4  & { $29.85\pm0.48$} &   { $4.8\pm0.21$} & {  \boldm $4.04\pm0.39$} \\
			WSC09-5  & { $6.14\pm0.19$}& { $4.65\pm0.36$}   & { \boldm $4.15\pm0.12$}  \\
			\hline
			
		\end{tabular}%
	\end{center}
	
\end{table}

\begin{table}
	\centering
	\caption{Average improvements over 30 runs, made by each local search technique in MAType-II.}
	\label{tab:improvements}
	\begin{center}

		\begin{tabular}{|l|l|l|l|}

			\hline
				\multicolumn{1}{|p{1.2cm}|}{\centering Task}&
			\multicolumn{1}{|p{1.5cm}|}{\centering Local Search Type-I} & 
			\multicolumn{1}{|p{1.5cm}|}{\centering Local Search Type-II} &  \multicolumn{1}{|p{1.7cm}|}{\centering No Improvements }  \\

			\hline

			WSC08-1  &{ $4\%$}  &{ $47\%$} &{ $49\%$} \\
			
			WSC08-2  &{ $0\%$}&{ $50\%$}&{ $50\%$} \\
			
			WSC08-3  &{ $3\%$}&{ $51\%$}&{ $46\%$} \\
			
			WSC08-4  &{ $2\%$}&{ $39\%$}&{ $59\%$} \\
			
			WSC08-5   &{ $4\%$}&{ $52\%$}&{ $44\%$}	\\
			
			WSC08-6  &{ $1\%$}&{ $52\%$}&{ $47\%$} \\
			
			WSC08-7  &{ $3\%$}&{ $35\%$}&{ $62\%$}   \\
			
			WSC08-8 &{ $1\%$}&{ $44\%$}&{ $55\%$} \\
			
			WSC09-1 &{ $3\%$}&{ $40\%$}&{ $57\%$}\\
			
			WSC09-2  &{ $1\%$}&{ $39\%$}&{ $60\%$} \\
			
			WSC09-3 &{ $3\%$}&{ $56\%$}&{ $41\%$} \\
			
			WSC09-4  &{ $4\%$}&{ $42\%$}&{ $54\%$} \\
			
			WSC09-5  &{ $1\%$}&{ $47\%$}&{ $52\%$} \\
			\hline
			
		\end{tabular}%
	\end{center}
	
\end{table}
\subsection{Discussion}\label{discussion}

Our experimental evaluations in Tables \ref{tab:results} and \ref{tab:time} show that the proposed method (MA-DWSC(II))which utilises a flexible local search operator and considers a communication link and its services, produces superior results than other techniques. 
Additionally, the effectiveness of the two local search strategies (Type-I and Type-I) in MA-DWSC(II) have been compared in terms of the number of improvements they made,  in Table \ref{tab:improvements}. As it is clearly demonstrated, Type-II has made significantly more improvements than Type-I. This verifies that it has been able to produce enhanced neighbours for the majority of the time. For example, for Task 08-1, in 47{\% } of local search recalls improvements have been made by neighbours which were produced by the local search Type-II, while only in 4{\% } the local search Type-I made improvements, and in 49{\% } of cases none of those neighbours outperformed the current solution. It does not restrict itself to include the Web service in the solution which is attached to the beginning of the longest communication link. The explanation to this is that if the Web service is located on a very distant place, including it in the composition can prevent the actual improvement in the solution. 

MA-DWSC(I) and MA-DWSC(II) utilise the domain knowledge in local search regarding the distribution and location of services. However, according to the results, MA-WSC outperforms MA-DWSC(I) in most cases. In other words, the neighbourhood size for local search in MA-WSC is large, which leads to a good performance of this local search technique at the cost of execution time. Task 09-3 has been used to exemplify all methods and their speed of convergence where the fitness is illustrated across generations (in seconds) in Fig. \ref{graph}.

An example solution for task 08-2 before and after undergoing local search Type-II is illustrated in Fig. 5. The Web service $f$ was replaced by the combination of two services, $m$ and $n$. This verifies that the link between $f$ and $g$ in Fig. 5(a) have been identified as the longest communication link in the solution. Although the number of services has become larger in Fig. 5(b) after applying local search, their total communication cost and time were less than the longest communication link ($f$ and $g$). Therefore, this local search has led to an improvement in the fitness value after being applied to the solution. 
\begin{figure}
	
	\label{fig:exampleofls}

	\centering
	\begin{tabular}{@{}c@{}}
		\includegraphics[width=4.25cm,height=2.5cm,frame]{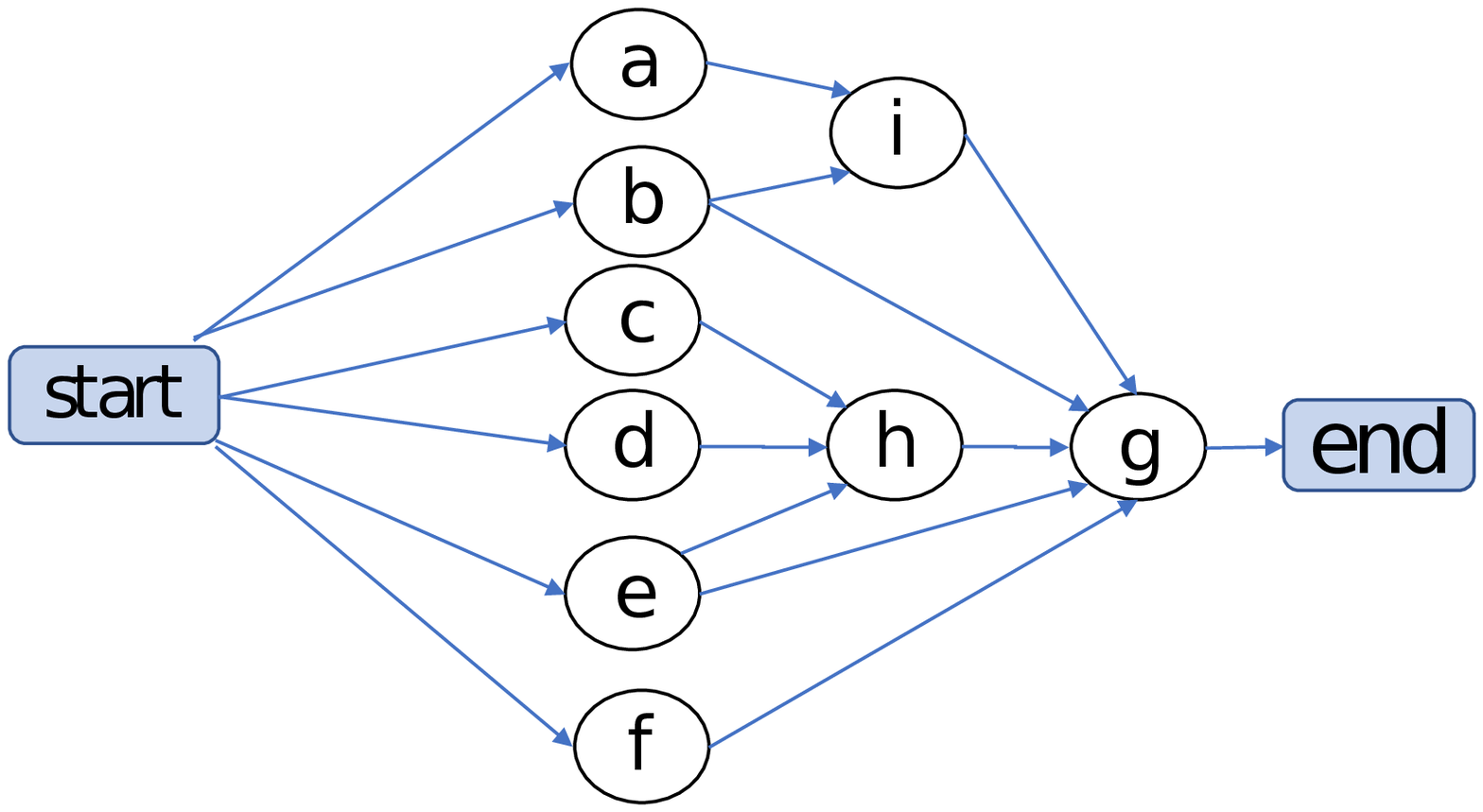}\\[\abovecaptionskip]
		\label{fig:com}
		
		
		\small  (a) 
		
	\end{tabular}
	\begin{tabular}	{@{}c@{}}
		\includegraphics[width=4.25cm,height=2.5cm,frame]{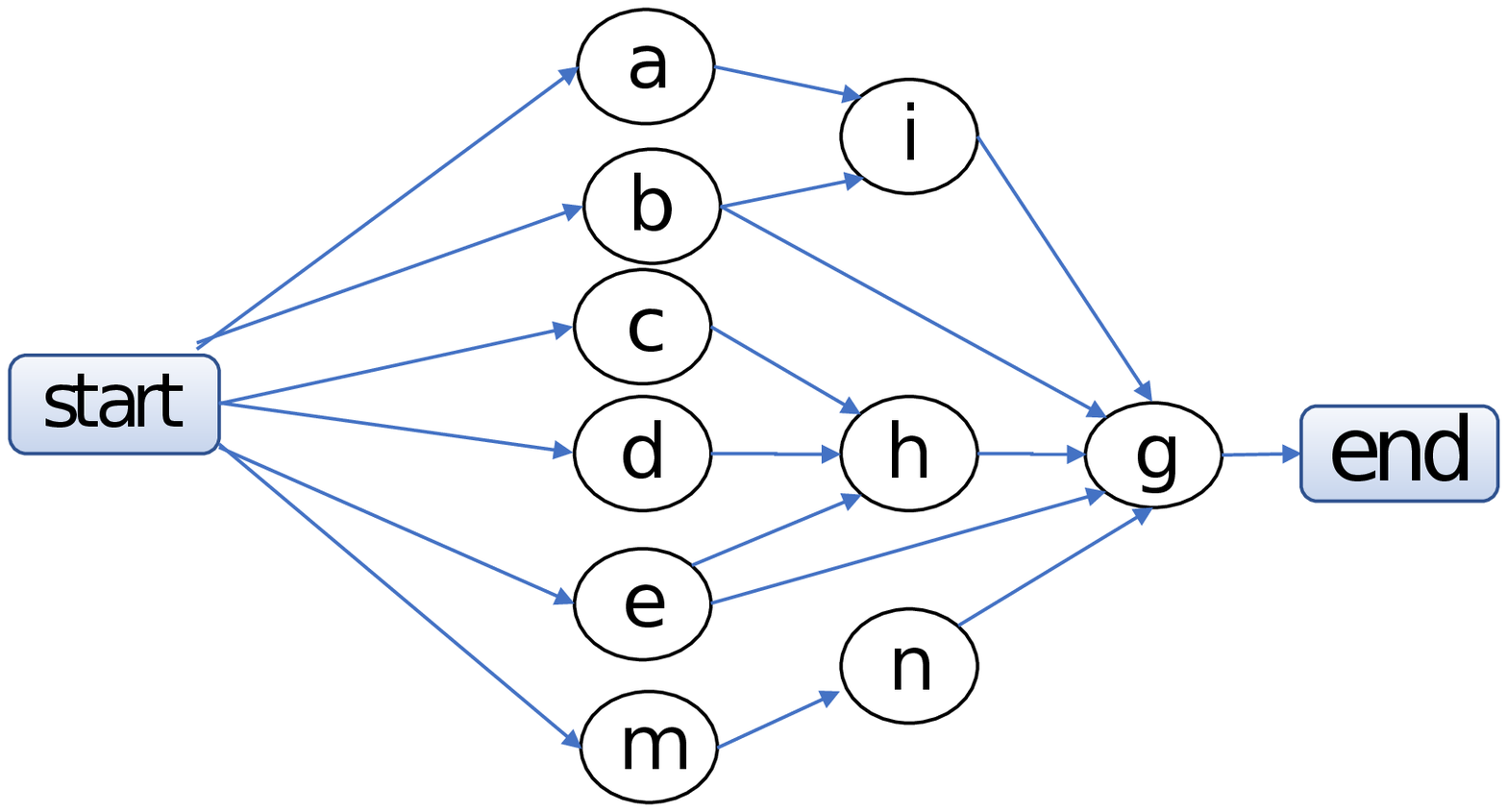}\\[\abovecaptionskip]
		\captionsetup{justification=centering}
		\label{fig:figls}
		\small (b)
	\end{tabular}

	

	
	
		\label{fig:lstype2}
	\caption{ An example of a solution (a) before (b) after,  applying local search Type-II.  }
\end{figure}

\begin{figure}
	\includegraphics[width=0.5\textwidth, height=5.5cm]{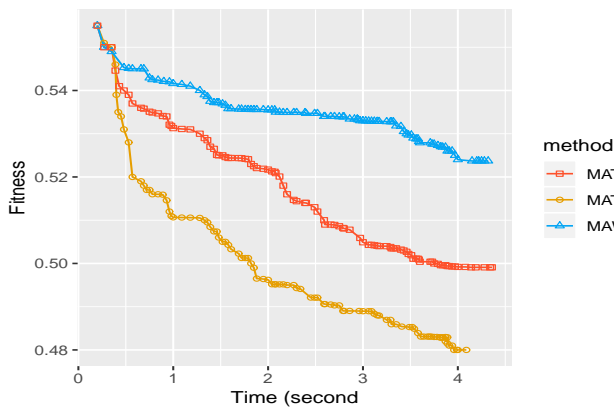}
	\label{convergencyspeed}
	\caption{Mean solution's fitness over time obtained for task 09-3 (Note: the smaller the IGD the better)}
\label{graph}
\end{figure}
\section{Conclusions}\label{conclusion}

In this paper, we studied the problem of the fully-automated composition of distributed data-intensive Web services. We have modelled the problem by considering distributed locations of data and services. We then proposed a memetic algorithm that hybridises GA with a novel local search method, using the location information of services and data. Our proposed memetic algorithm (MA-DWSC(II)) was based on a variable-length GA and a local search. The local search was designed using domain knowledge of the DWSC and took into account replacing a communication link along with its services with another Web service in the repository. Our experimental evaluation using benchmark datasets showed that MA-DWSC(II) outperformed two other methods, i.e., MA-WSC and MA-DWSC(I). This high performance is presumably due to its higher flexibility which allows a path replacement in the composition graph. To provide the same condition, all those techniques were using LCS crossover operator. This crossover has shown to be extremely effective for distributed DWSC \cite{Sadeghiram2019Distance}, which helps the offspring to inherit good parts from parents. In the future, we will expand our study to further improve our proposed algorithm.

%
%
%
%

\end{document}